\documentclass[conference,compsoc]{IEEEtran}

\ifCLASSOPTIONcompsoc
  \usepackage[nocompress]{cite}
\else
  \usepackage{cite}
\fi
\usepackage{url}
\usepackage{float}
\usepackage{makecell}

\ifCLASSINFOpdf
\else
\fi
\usepackage{amsmath}
\usepackage{todonotes}
\usepackage{tabularx, booktabs}

\usepackage{amssymb}
\usepackage{amstext}
\usepackage{amsmath}
\usepackage{bm}

\usepackage{fancyhdr}

\fancypagestyle{firstpage}{%
  \lhead{
   \footnotesize{This paper is under review at the journal of Computer Vision and Image Understanding}
}}

\begin{document}
%
\title{LOST: A flexible framework for semi-automatic image annotation}



%
\author{\IEEEauthorblockN{
Jonas J\"ager\IEEEauthorrefmark{1}\IEEEauthorrefmark{2}\IEEEauthorrefmark{3},
Gereon Reus \IEEEauthorrefmark{1}\IEEEauthorrefmark{3},
Joachim Denzler \IEEEauthorrefmark{2},
Viviane Wolff\IEEEauthorrefmark{1} and
Klaus Fricke-Neuderth\IEEEauthorrefmark{1}}
\IEEEauthorblockA{\IEEEauthorrefmark{1}Department of Electrical Engineering and Information Technology,
Fulda University of Applied Sciences, Germany\\ 
Email: jonas.jaeger@et.hs-fulda.de}
\IEEEauthorblockA{\IEEEauthorrefmark{2}Computer Vision Group, 
	Friedrich Schiller University Jena, Germany\\
	www.inf-cv.uni-jena.de}
\IEEEauthorblockA{\IEEEauthorrefmark{3}L3P UG, Germany\\
	www.lost.training}
}


\maketitle
\thispagestyle{firstpage}

\begin{abstract}
State-of-the-art computer vision approaches rely on huge amounts of annotated data. 
The collection of such data is a time consuming process since it is mainly performed by humans.
The literature shows that semi-automatic annotation approaches can significantly speed up the annotation process 
by the automatic generation of annotation proposals to support the annotator.
In this paper we present a framework that allows for a quick and flexible design of  semi-automatic annotation pipelines.
We show that a good design of the process will speed up the collection of annotations.
Our contribution is a new approach to image annotation that allows for the combination of different annotation tools and machine learning algorithms in one process. 
We further present potential applications of our approach.  
The source code of our framework called LOST (Label Objects and Save Time)
is available at: \url{https://github.com/l3p-cv/lost}.
\end{abstract}


%
\IEEEpeerreviewmaketitle

\section{Introduction}
A huge amount of annotated data is the key to success in machine learning and computer vision.
However the annotation process is still extremely elaborate, 
since humans or even experts in a specific field are required.
Therefore a good annotation tool and smart annotation strategies are essential 
to build large datasets of sufficient quality.

In recent years the community focused on three main points to save time 
and improve annotation quality while collecting datasets for computer 
vision research. 

1) Crowdsourcing
approaches as presented in \cite{Su2012, Kovashka2016} 
have been utilized to collect huge amounts of annotations via crowdsourcing 
platforms such as Amazon Mechanical Turk.
With this strategy the overall time for dataset collection is 
reduced significantly by employing a large number of annotators.

2) A second focus of the community was to optimize the annotation 
process itself by supporting the human annotator.
The main idea here is to reduce the human interaction with the 
annotation tool to save time,
while maintaining the quality of the annotations 
\cite{Russakovsky2015, Papadopoulos2016, Papadopoulos2017, Papadopoulos2017a, Acuna2018}.


3) The third main focus was on the development of annotation tools and their user interfaces, 
since the user experience with such a tool is important for the annotators motivation and the quality of the annotations.
A wide variety of annotation tools is described in the literature \cite{Russell2008, Vondrick2010, Bianco2015, Dias2019}.

Our contribution is a flexible pipeline concept to model semi-automatic image 
and video annotation.
This approach allows to combine multiple annotation tools and machine learning algorithms as one process in a
building block style.
We visualize the whole annotation process in a web-based user interface.
Furthermore,
we provide an annotation interface to assign labels to clusters of images or annotations e.g. boxes (see Section~\ref{sec:mia}).

The open source implementation called LOST (Label Objects and Save Time) 
is available on GitHub (\url{https://github.com/l3p-cv/lost}). 
This implementation contains an annotation process visualization,
two annotation tools, a tree-based label management and an annotator management.
Our tool allows researchers to design and run their own annotation pipelines in a quick and consistent way.
Furthermore each developer can inject his own Python scripts to gain full control over the process.
A single instance of LOST can be easily set up with docker,
to be used as stand alone application on a single machine.
LOST can also be set up as a cloud application to allow collaborative annotation via the web.
If required, LOST is able to distribute computational workload across multiple machines.

\section{Related Work}

\subsection{Approaches to Support the Annotator}
The authors of \cite{Papadopoulos2017, Papadopoulos2017a} 
use point annotations to train object detection models.
Papadopoulus et al. \cite{Papadopoulos2017a} 
report a speed up of the total annotation time by a factor of 9 to 18
compared to traditional bounding box annotations.
The performance of their detectors is close to a detector trained with hand drawn bounding box annotations.
Russakovsky et al. \cite{Russakovsky2015} 
use point supervision to create a segmentation model that is more accurate than models trained with full supervision given a fixed time budget.

The authors of \cite{Papadopoulos2016} 
generate bounding box proposals to ask the annotators if a box is correct or not. 
A bounding box is considered as correct if the intersection over union \cite{Everingham2015}
with a tight box around the object is greater than $0.5$.
After the annotators verification step, 
the object detector is retrained with the new boxes.
This approach reduces the human annotation time by a factor of $6$ to $9$,
while achieving a mAP of $58\%$ on Pascal VOC 2007.
When training with full supervision the authors achieve a mAP of $66\%$.
In \cite{Konyushkova2018} 
an agent is trained to select the best strategy for bounding box annotation. 
Two strategies are considered to be selected by the agent: 
either box verification \cite{Papadopoulos2016} 
as described above or manual box drawing.


\subsection{Annotation Tools/ Interfaces}
Russel et al. \cite{Russell2008} 
present a general purpose web-based image annotation tool called \textit{LabelMe} to create polygon annotations.
Vondrick et al. \cite{Vondrick2010} 
propose a web-based tool called \textit{VATIC}
for semi-automatic video annotation in a crowdsourcing setup.
They use bounding box annotations and link these boxes to create ground-truth tracks
for video sequences
within an optimized user interface.
Similar to \textit{VATIC}, 
\textit{iVAT} \cite{Bianco2015} 
and \textit{ViTBAT} \cite{Biresaw2016}
are tools for semi-automatic video annotation. 
In contrast to \textit{VATIC} these tools are not web-based.	
Polygon-RNN \cite{Acuna2018} 
follows an interactive approach for faster polygon annotation.
A recurrent neural network is utilized to support the human annotator by iteratively sending annotation proposals to the user interface.
Qin et al. \cite{Qin2018} present the semi-automatic tool \textit{ByLabel}
that supports the annotator in segmentation tasks.
The tool \textit{FreeLabel} \cite{Dias2019} is also designed to collect segmentation masks.
It uses scribble annotations as seeds for the region growing
algorithm to create a semi-automatic segmentation result.

BIIGLE \cite{Langenkaemper2017} 
is a web-based tool that is especially designed for the annotation and exploration of marine image collections.
It provides different annotation and review interfaces,
a project management and a user management.
In difference to the other tools it also implements a label tree management.
In contrast to LOST, 
Biigle does not implement a flexible pipeline system.

Each of the above mentioned annotation tools was designed with a specific application and annotation process in mind.
Due to that, 
these annotation tools have hard-coded a specific process/ algorithm and model only a single use case.
This leads to the fact that every time a new annotation approach was tested a new tool was implemented.
In contrast to that, 
our framework is able to model multiple semi-automatic annotation approaches, e.g. \cite{Russakovsky2015, Papadopoulos2016, Papadopoulos2017, Papadopoulos2017a},
in a consistent and fast way.
In this sense our proposed framework is a generalization of single purpose annotation tools.

\section{Approach}
We propose a framework for semi-automatic image annotation.
This framework allows any combination of machine learning algorithms and 
annotation interfaces in a building block style.

\begin{figure}[t]
	\begin{center}
		\includegraphics[width=1\linewidth]{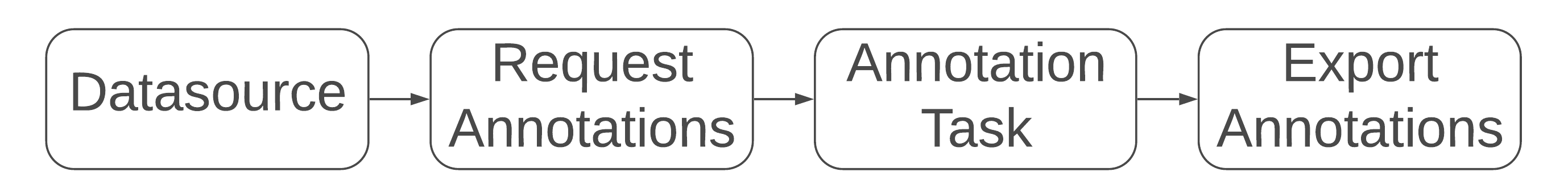}
	\end{center}
	\caption{Schematic illustration of an example annotation pipeline. 
		A \textit{datasource} represents a set of images that should be annotated e.g. the Pascal VOC dataset. 
		The next element in the pipeline is a \textit{script} 
		that requests annotations for the connected \textit{annotation task}.
		When the \textit{annotation task} was processed by a human annotator for all requested images 
		another Python \textit{script}
		is used to export all annotations to a csv file.
		After that the csv file can be downloaded in the web GUI.
	}
	\label{fig:single-stage}
\end{figure}


Figure~\ref{fig:single-stage}
represents a simple example, 
while much more complex processes can be modeled.
In general an annotation process is defined as a directed graph.
Each node in a graph represents one building block and the connections between the nodes
define the order in which the building blocks are processed.
Also information, 
such as annotations,
can be exchanged between connected elements and accessed via the framework API.

The basic building blocks of an annotation pipeline are \textit{datasources}, 
\textit{annotation tasks}
and \textit{scripts} that implement different algorithms.
By means of these building blocks,
annotation tasks for humans can be combined with 
machine learning and other algorithms 
in a flexible way.



\subsection{Flexibility}

The main feature of LOST is a flexible pipeline concept.
An example for its flexibility is the combination of different annotation interfaces in one pipeline.
When combining a single image annotation interface (SIA, Section~\ref{sec:sia})  
with a multi image annotation interface (MIA, Section~\ref{sec:mia}),
annotation tasks can be split into object localization and class label assignment.
See Section~\ref{sec:two-stage-annotation} for a detailed description of a two-stage annotation approach where SIA is used to draw bounding boxes and MIA to assign a class label to each box.

Flexible pipelines allow also to combine any kind of machine learning algorithm with an annotation interface to realize semi-automatic annotation approaches.
For example, 
SIA can be combined with a script that implements an object detector that generates bounding box proposals. 
MIA can be connected to a script element that implements an algorithm that clusters images based on their visual similarity in order to speed up the annotation.
See Section~\ref{sec:two-stage-loop} 
for a semi-automatic annotation pipeline that combines SIA and MIA with machine learning algorithms.

LOST allows also to model iterative annotation processes when adding loop elements 
to a pipeline (see Section~\ref{sec:two-stage-loop}).
In this way lifelong learning \cite{Chen2013, Kaeding2016} 
and active learning \cite{Settles2009, Brust2019} 
approaches can be realized with LOST.

\subsection{Building Blocks}
An annotation pipeline (annotation process) 
can be composed of six different building block types.
These are \textit{datasources}, 
\textit{scripts}, 
\textit{annotation tasks}, 
\textit{loops},
\textit{data exports} and \textit{visualizations}.
After a pipeline was designed as a composition of the different building blocks,
it can be loaded into the LOST framework. 
When starting (instantiating) a pipeline , 
each element can be parameterized.
For example, for a \textit{datasource} 
a set of images will be selected.
Another common example is the selection of a user or group that will perform an annotation task
to parameterize an \textit{annotation task}-element in a pipeline.

A \textit{datasource} 
represents a set of images or videos that can be used by connected elements in
the pipeline, 
for example by one or many \textit{scripts}.
Such a \textit{script}-element 
is an arbitrary algorithm implemented in Python that communicates with connected elements via the framework API.
The main purpose of a \textit{script}
is to generate object proposals or to cluster images for semi-automatic annotation.
An example of such an object proposal could be a bounding box generated by a RetinaNet \cite{Lin2017a} object detector.
When a \textit{script} has generated proposals for all images that should be annotated
it will send them to an \textit{annotation task}, 
where the proposals will be displayed to a human annotator. 

An \textit{annotation task}-element links users, annotation tools and labels.
There are two types of annotation tools that can be used in our current implementation.
The first tool was designed to annotate single images (Section~\ref{sec:sia}) and with the second tool clusters of images can be annotated (Section~\ref{sec:mia}). 
Labels are represented as trees (Section~\ref{sec:label}).
In this way we are able to model label hierarchies.

\textit{Loop}-blocks can be used for iterative annotation processes
where parts of a pipeline need to be executed multiple times until a certain criterion is fulfilled.
\textit{Loops} are often useful to model active learning or continuous learning approaches.
See Section~\ref{sec:two-stage-loop} 
for an example how a \textit{loop} 
can be used in a pipeline.

The last two element types that can be used are \textit{data exports} 
and \textit{visualization}-elements.
\textit{Data exports} 
are used to provide any files created by \textit{scripts} 
via the GUI for download.
Similar to \textit{data exports}, 
\textit{visualizations} display images created by \textit{scripts}
within the web interface.

\subsection{Multi Image Annotation (MIA) Interface} \label{sec:mia}
MIA serves to assign labels to clusters of images or annotations. 
The main idea is that visual similar objects are likely to get the same label.
This idea is related to the cluster-based approach to fish annotation proposed by \cite{Boom2012a}.

In an ideal world a cluster contains only images that belong to the same class. 
A human annotator has the task of sorting out images that do not belong to the cluster.
Therefore, 
the same label can be assigned to large number of images at the same time. 
In the same way labels can be assigned to point, 
box, 
line and polygon annotations.

LOST provides the first open source implementation of a MIA interface.
Due to the pipeline concept,
MIA can be combined with a SIA interface in one annotation process (see Section~\ref{sec:two-stage-annotation}). 
In this way, class labels and object localizations can be annotated in specialized interfaces for each task to speed up annotation.
Furthermore, MIA can be combined with different cluster algorithms by connecting it to a \textit{script} element that will sort annotations or images into clusters.


\subsection{Single Image Annotation (SIA) Interface} \label{sec:sia}
SIA is designed to create \textit{polygons, points, lines} 
and \textit{bounding box} 
annotations. 
To each annotation and the whole image,
a class label can be assigned.
Also the assignment of multiple class labels is possible.
Furthermore, 
the tool is configurable to allow or deny different types of user actions and annotations depending on the use case. 
For example, 
you can specify that only class label assignment is possible, 
but no other modifications are allowed.
When combining SIA with an algorithm that generates annotation proposals semi-automatic annotation can be realized.
See Section~\ref{sec:two-stage-loop} for an example.


\subsection{Label Management} \label{sec:label}
Labels are managed in label trees to model label hierarchies.
Multiple label trees can be created and edited. 
When starting an \textit{annotation task} 
a whole label tree or a composition of subtrees can be selected as possible labels.
During the \textit{annotation task} 
the annotator can assign one of the possible labels to an annotation.

\subsection{Comparison to State-of-the-Art Annotation Tools}
Table~\ref{tab:comparison} presents an overview of the key ideas of LOST in comparison to 
other open source tools.
LOST is the only tool with a flexible pipeline system, 
where multiple annotation interfaces and algorithms can be combined in one process.
There are many tools that where build on web technologies to enable collaborative annotation.

Table~\ref{tab:comparison} 
shows also three different annotation interface designs 
and if they are implemented for a specific tool:

1) SIA interfaces are used to annotate single images with different annotations like 
\textit{points, boxes, lines, polygons, etc.}
Most tools do focus on a SIA interface.

2) MIA interfaces are used to present clusters of similar images and assign a label to a whole cluster of images.
This idea of MIA was described by Boom et al. \cite{Boom2012a}.
In the presented comparison LOST is the only tool that has implemented this type of interface.

3)Image sequence annotation (ISA) interfaces are especially designed to annotate video sequences with tracking information.
In most cases these interfaces are extensions of a SIA interface.
In the current version, LOST has no ISA interface to annotate tracks.
The implementation of ISA is planned for the near future.

Another difference between LOST and the other tools is that user-defined python scripts can be executed as part of an annotation pipeline.
In such scripts any python code can be implemented and information with other pipeline elements can be exchanged to get full control over the process. 

\begin{table*}[!t]
	\caption{\label{tab:comparison}LOST features in comparison to other open source annotation tools.}
	\centering
	\begin{tabular}{|p{3.2cm}|p{2cm}|p{2cm}|l|l|l|p{2.2cm}|l|}
		\hline
		Features & Flexible Pipelines & Collaborative Annotation & SIA$^{1}$ & MIA$^{2}$ & ISA$^{3}$ & Run User defined Scripts & License\\
		\hline
		\hline
		\textbf{LOST} & yes & yes & yes & yes & -- & yes & MIT \\
		\hline
		VoTT \cite{VoTT}& -- & -- & yes & -- & -- & --& MIT \\
		\hline
		CVAT \cite{CVAT} & -- & yes & yes & -- & yes & -- & MIT\\
		\hline
		Mask Editor \cite{Zhang2018} & -- & -- & yes & --& -- & -- & Free\\
		\hline
		FreeLabel \cite{Dias2019} & -- & yes & yes & -- & -- & -- & NPOSL-3.0 \\
		\hline
		Polygon-RNN++ \cite{Acuna2018} & -- & n/a & yes & -- & -- & -- & n/a \\
		\hline
		ByLabel \cite{Qin2018} & -- & -- & yes & -- & -- & -- & GPL-3.0 \\
		\hline
		LabelMe \cite{Russell2008} & -- & yes & yes & -- & -- & -- & MIT \\
		\hline
		VATIC \cite{Vondrick2010} & -- & yes & -- & -- & yes & -- & MIT \\
		\hline
		iVAT \cite{Bianco2015} & -- & yes & -- & -- & yes & -- & n/a \\
		\hline
		ViTBAT \cite{Biresaw2016} & -- & -- & -- & -- & yes & -- & Free \\
		\hline
		VIA \cite{Dutta2016} & -- & -- & yes & -- & yes & -- & BSD-2 \\
		\hline
		\multicolumn{8}{@{}l}{$^1$Single Image Annotation Tool, 
			$^{2}$ Multi Image Annotation Tool, $^{3}$ Image Sequence Annotation Tool}
	\end{tabular}
\end{table*}

\section{Case Studies}
In this Section we show how annotation pipelines can be modeled and executed within our web-based framework.

All case studies are performed and analyzed on the Pascal VOC2012 \cite{Everingham2015} dataset.
We use $80\%$ of the VOC validation split to select the images for annotation and $20\%$ for evaluation of trained models.
The VOC training split is used for initial training.
For our experiments we utilize the keras implementations of the respective models. The experiments are executed inside a NVIDIA Docker container integrated in LOST. The server is equipped with a NVIDIA Geforce GTX1080Ti, an Intel i7-8700K and 16GB DDR4 RAM.

\subsection{Single-Stage Annotation} \label{sec:single-stage}
In our first experiment we create a baseline for simple bounding box annotation in the proposed framework as it is possible in most annotation tools.
In order to do that, 
we use 200 images randomly selected from the Pascal VOC validation set.
In this selection we guarantee that each class is present in at least 10 images.
Two annotators perform bounding boxes annotation according to the VOC annotation guidelines on each image.
Box drawing and class assignment is performed with the SIA tool (Section~\ref{sec:sia}).


\paragraph{Results}

\textit{Annotator1} achieved a mAP of $81\%$ 
and \textit{Annotator2} a mAP of $82\%$ compared to the Pascal VOC ground truth annotations,
when using an intersection over union threshold of $0.5$.
Among each other the annotators agreed on $86\%$ of the annotated boxes.
This show that even among human annotators the level of agreement was below $87\%$ mAP in our experiment.
Which is an interesting fact when considering that object detectors are trained with human data.

On average the annotators needed $11.15$ seconds
to draw a bounding box and assign a class label to this box.
The average time to annotate an image was $28.2$ seconds.
For the annotation of $200$ images $102.75$ minutes of annotation time was required.


\subsection{Two-Stage Annotation} 
\label{sec:two-stage-annotation}
In this experiment we combine the single image annotation (SIA)
with the multi image annotation tool (MIA) 
in a two-stage annotation process. 
In the first stage the annotator has the task to draw bounding boxes according to the VOC annotation guidelines.
In this stage no class labels will be assigned.
In the second stage all boxes from the first stage will be clustered according to their visual similarity and presented in the MIA interface.
The annotators task here is to assign a label to the whole cluster. 
If an image does not belong to that cluster, 
it should be removed by the annotator.
For example, 
if there are 19 persons and one cat in the view, 
the annotator should remove the cat and select the label person for the remaining images.


For this experiment, 
a similar setup as in Section~\ref{sec:single-stage} is used.
The same 200 images from the VOC validation set are annotated by a human annotator,
while this time two annotation stages are performed.

We test two clustering strategies. 
First we extract CNN features 
for all annotated boxes from stage one 
and use the K-Means algorithm to cluster all images.
For feature extraction we use the last pooling layer of \textit{ResNet50} \cite{He2016} 
that was pretrained on ImageNet \cite{Deng2009}.

As second clustering approach we fine-tune \textit{ResNet50} 
on a small subset of the Pascal VOC training set and utilize the networks predictions directly for clustering.

\begin{figure}[t]
	\begin{center}
		\includegraphics[width=1\linewidth]{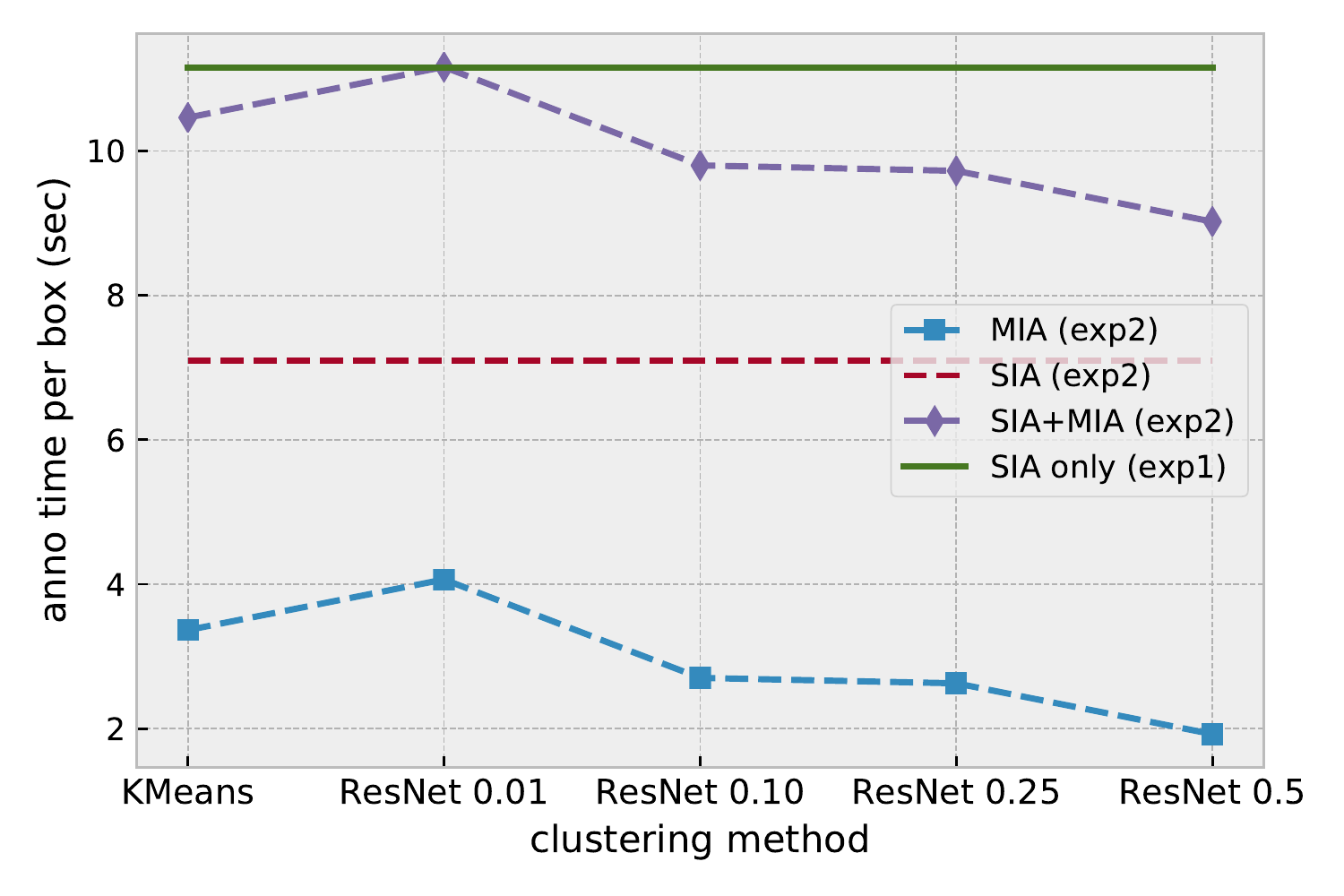}
	\end{center}
	\caption{Average times for bounding box annotation with a single-stage approach in comparison to a two-stage approach. 
		\textit{SIA only (exp1)} indicates the time for the single-stage approach where 
		class label assignment and box drawing was performed in one step.
		\textit{SIA+MIA (exp2)} shows the times for the two-stage approach for different 
		clustering algorithm setups.
		\textit{SIA (exp2)} and \textit{MIA (exp2)} indicate the times for stage one and
		stage two of our two-stage approach.
	}
	\label{fig:exp2-times}
\end{figure}

\paragraph{Results}
In the first stage the annotator used the SIA annotation tool and needed $7.1$ 
seconds (see Figure~\ref{fig:exp2-times}, \textit{SIA~exp2}) to draw a bounding box and $18.9$ 
seconds to draw boxes for all objects in an image.
On average he annotated $2.7$ 
boxes per image,
while the total annotation time for 200 images was $63.8$ minutes in the first annotation stage.

Figure~\ref{fig:exp2-times} 
shows also the average annotation time per box for class label assignment in the second annotation stage (\textit{MIA~exp2}).
The x-axis indicates the different cluster methods, 
where K-Means and ResNet50 were used.
ResNet50 was fine-tuned with $1\%=57$ images, $10\%=571$ images, $25\%=1429$ images 
and $50\%=2858$ images 
of the Pascal VOC training split.
We see that the two-stage annotation process (Figure~\ref{fig:exp2-times}, \textit{SIA+MIA~exp2}) is faster than single stage annotation if the clustering algorithm works well. 
Only when ResNet50 was fine-tuned with $1\%$ of the training data, 
the two-stage process consumed the same time as one-stage.
The mAP of the created boxes compared to the VOC ground-truth data was almost equal and around $80\%$ 
for all approaches.
With the fastest two-stage approach it took $81.1$ 
minutes to annotate $200$ images, where single-stage annotation took $102.75$ minutes.

When considering that class label assignment plus box drawing with SIA needs $11.15$ seconds 
and box drawing only needs $7.1$s,
we know that pure class label assignment with SIA takes $4.05$ seconds.
The fastest annotation approach with MIA takes $1.92$ seconds per box, 
which is a speed up in class label assignment by a factor of two.


\paragraph{Use cases}
The main idea here was to break down the complex task of simultaneous bounding box drawing and class label assignment into two separate tasks that are simpler, 
while supporting the annotator with a preclustering during class label assignment.
In this way it was also possible to split the annotation work into a simple and an expert task,
where in most cases expert knowledge is required for class label assignment as in many biological \cite{Boom2012a} or medical applications \cite{Esteva2017}.
We saw that when using the MIA annotation tool and a good clustering algorithm the time for class label assignment was reduced by a factor of two compared to label assignment with SIA.
This allows for saving expensive annotation time for expert tasks.
For the experts,
who will not need to draw a box,
annotation time is reduced by a factor of $3$ to $6$. 

\subsection{Two-Stage Annotation in the loop} \label{sec:two-stage-loop}
In this experiment we show how to implement an iterative annotation process within our framework.
As in Section~\ref{sec:two-stage-annotation} 
we model a two-stage annotation pipeline composed of a single image annotation task and 
a clustered image annotation view.
In difference to the previous experiment we use semi-automatic support in both annotation stages, 
do not use any VOC data for pretraining and put everything into a loop.
In other words we assume that we have no annotated data in the beginning and try to get a better automatic support for the annotator over time.

As in the previous experiments we use images from the Pascal VOC2012 validation dataset for annotation.
Due to the iterative setup we use $150$ images per iteration that will be processed by two human annotators.
In contrast to the experiment in Section~\ref{sec:single-stage},
both annotators work on the same annotation task to split the workload.


\begin{figure}[t]
	\begin{center}
		\includegraphics[width=0.8\linewidth]{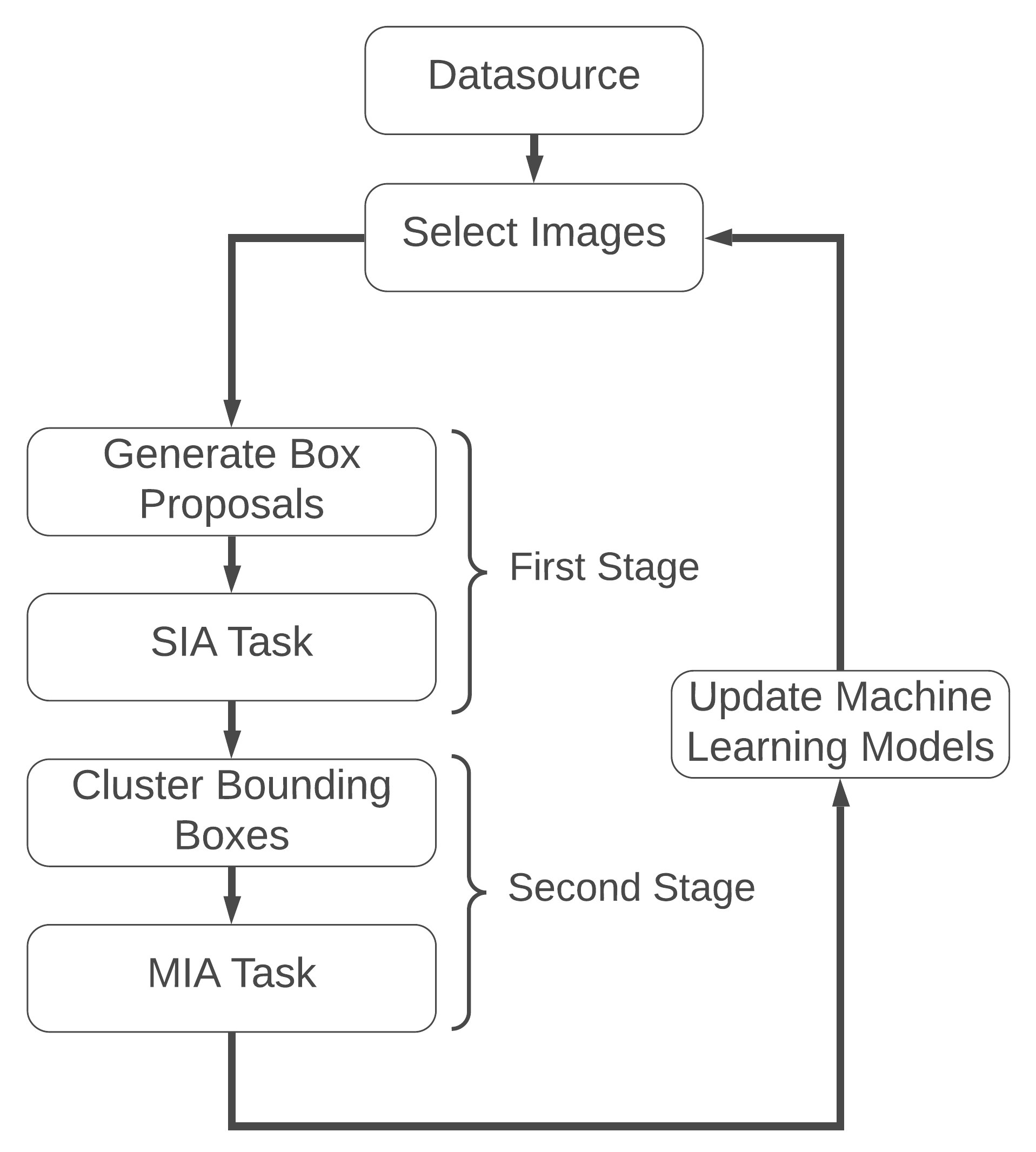}
	\end{center}
	\caption{Schematic illustration of two-stage annotation in the loop,
		as it is modeled in the proposed framework.
		The \textit{datasource} 
		provides the dataset and a Python \textit{script}
		selects images that should be annotated in stage one.
		Then bounding box proposals are generated for a
		\textit{single image annotation task}
		in order to support the human annotator when drawing boxes.
		In stage two, 
		a \textit{script} 
		clusters the bounding boxes by visual similarity.
		After that, 
		\textit{MIA} is used to assign class labels to clusters of 
		similar boxes.
		In the last step 
		all machine learning models will be retrained 
		with annotations from previous annotation tasks.
		When the training was performed, 
		a new iteration starts.
	}
	\label{fig:two-stage-loop}
\end{figure}

Figure~\ref{fig:two-stage-loop} 
shows a high level view of the annotation pipeline.
In the first annotation stage RetinaNet \cite{Lin2017a} 
is used for bounding box proposal generation in order to support the human annotator.
As proposals we use all boxes with a confidence value above $0.4$.
After each iteration, 
RetinaNet will be retrained with all annotations from previous iterations.
When all images have been processed by RetinaNet a SIA 
task will be performed by the human annotators.
Since there are no annotations in the first iteration, 
no box proposals are generated in the first iteration.
The annotators are instructed to draw bounding boxes around all VOC2012 objects in the images.

In the second annotation stage ResNet50 \cite{He2016} 
is used to cluster all bounding boxes by class.
We use pretrained ImageNet weights for initialization and fine-tune ResNet50 after each iteration with all annotations from previous iterations.
In the first iteration where no class label annotations are available, 
we use ImageNet class predictions for clustering.
The idea here is that visual similar images will get the same class label, 
even if the predicted class is not part of Pascal VOC. 
After that, 
the MIA tool 
is used to correct the proposed clusters and to assign class labels to the clustered box annotations.
When the second stage was processed, 
the next loop iteration will start. 

\paragraph{Results}
\begin{figure}[t]
	\begin{center}
		\includegraphics[width=1\linewidth]{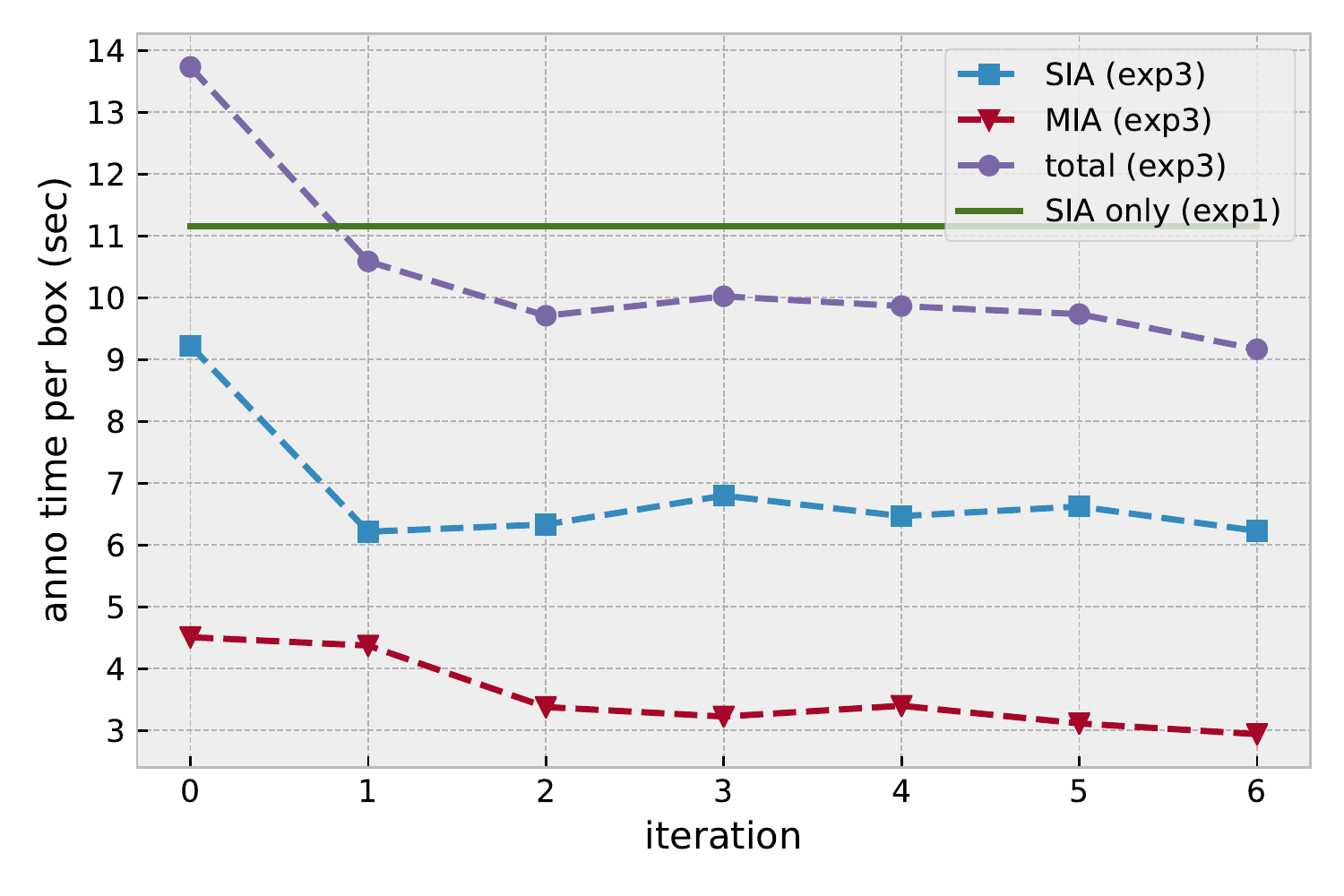}
	\end{center}
	\caption{Average annotation time per box for two-stage annotation in the loop (\textit{total exp3})
		vs. single-stage annotation (\textit{SIA only exp1}). }
	\label{fig:exp3-times}
\end{figure}
Figure~\ref{fig:exp3-times} 
presents the average annotation time per box per iteration.
We see times for box drawing in the first annotation stage (\textit{SIA exp3}),
the times for class label assignment in the second stage(\textit{MIA exp3}), 
the total time per box for the looped two-stage annotation approach (\textit{Total exp3})
and the single-stage annotation approach for comparison (\textit{SIA only exp1}).
In the first iteration, 
when no box proposals are generated and the cluster algorithm is not fine-tuned to the VOC dataset,
the looped two-stage approach is slower than single-stage annotation.
But in the following iterations when RetinaNet and ResNet50 are fine-tuned with the annotations of the previous iterations,
the looped two-stage annotation process gets faster than single-stage annotation.

\begin{figure}[t]
	\begin{center}
		\includegraphics[width=1\linewidth]{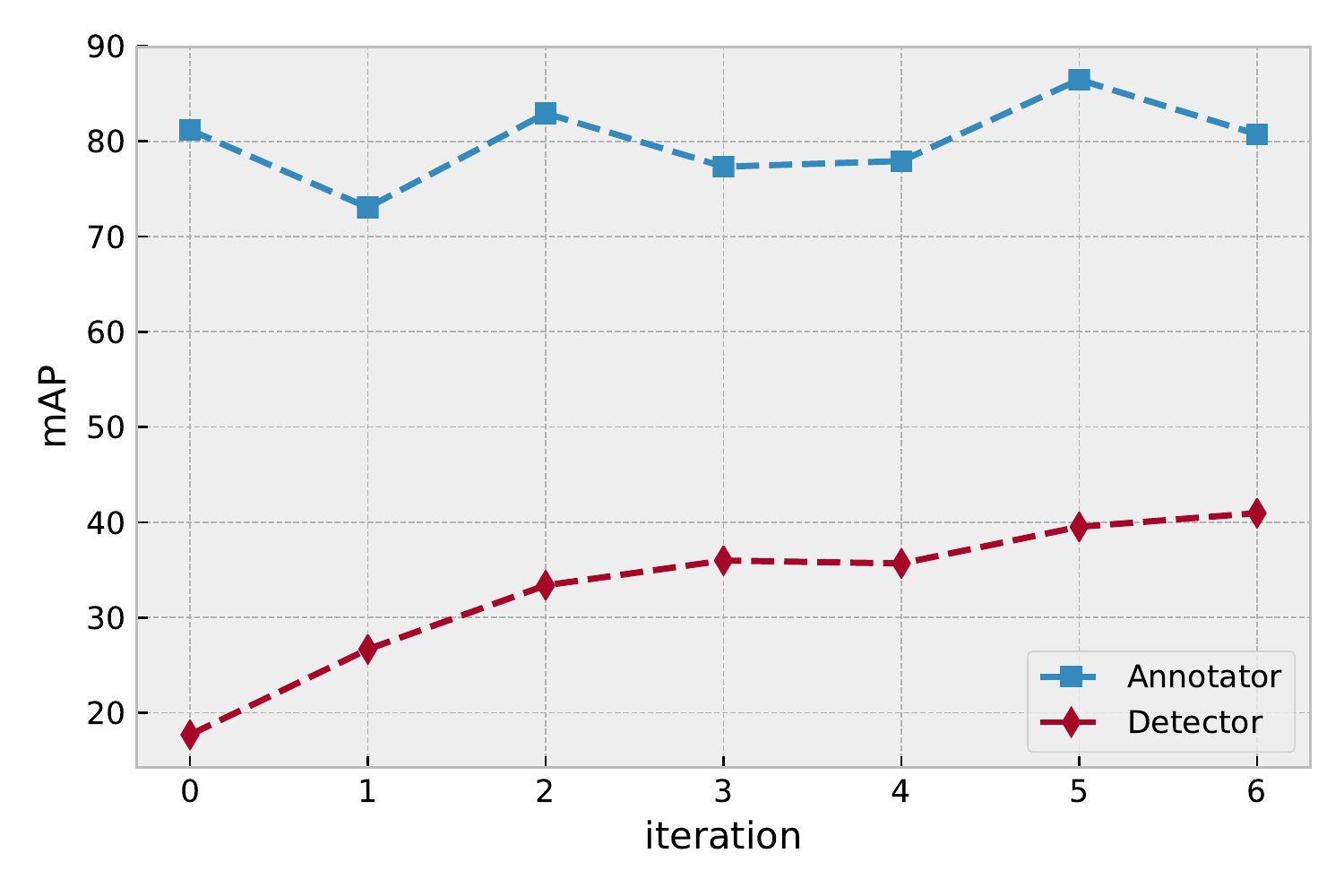}
	\end{center}
	\caption{Mean average precision per iteration for the \textit{two-stage in the loop}
		approach.
		The \textit{Annotator} 
		curve indicates the quality of the annotated boxes created by the human annotators compared to the ground truth boxes provided in the PascalVOC dataset.
		The \textit{Detector} 
		graph shows the mAP of the proposals generated by RetinaNet compared to the PascalVOC ground-truth annotations.
	}
	\label{fig:exp3-mAPs}
\end{figure}

Figure~\ref{fig:exp3-mAPs} 
shows the corresponding mAPs of the human-annotated boxes (\textit{Annotator}) 
and the detector performance of RetinaNet per iteration.
While the detector performance increases, 
the performance of the created annotations seems to be stable around a mean of $80\%$ mAP with a standard deviation of $4.3\%$. 
This deviation most likely reflects the annotators attention level and the difficulty of the images that have been annotated. 

\paragraph{Use cases}
We found that a looped two-stage annotation approach would be beneficial if there are no ground-truth data available in the beginning of the annotation process.
This approach creates annotations that have an equal quality compared to the single-stage approach
and trains a detector on-the-fly, 
while taking less annotator time than single-stage annotation.
Since it is modeled as two-stage process, 
the annotation work can be split in a simple and an expert task as in Section~\ref{sec:two-stage-annotation}.
It is notable that the training process of the machine learning models takes additional time compared to annotation approaches without machine learning elements,
but we found that this is no problem when performing the training in time slots when the annotator needs to rest anyway e. g. over night. 




\section{Conclusion}
To the best of our knowledge, we present the first framework for a flexible design and instantiation of image annotation pipelines.
Our approach enables the combination of different annotation tools and machine learning algorithms in one process.
We also provide an annotation interface called MIA (multi image annotation) to annotate whole clusters of images at the same time.

Our case studies show how our framework can be used to model machine learning based semi-automatic annotation pipelines and iterative annotation approaches.
In Section~\ref{sec:two-stage-annotation} 
we found that simple clustering in combination with the MIA annotation interface can speed up class label assignment by a factor of two compared to single stage annotation.
We also show that an annotation task can be split in an expert and a simple task,
which can significantly reduce expensive expert annotation time.
We further show in Section~\ref{sec:two-stage-loop}
that a looped two-stage approach is beneficial when no annotation data is available in the beginning. 
The quality of the created annotations is kept high while the time spent for box annotation gets smaller over time.

In future, we plan to release the missing image sequence annotation interface (ISA)
that is specialized to annotate tracks.
We also want to implement an interface to Mechanical Turk for crowdsourcing applications.

\section{Acknowledgments}
We thank Clemens-Alexander Brust and Christoph K\"ading for the helpful discussions on lifelong machine learning and Paul Bodesheim for his comments that greatly improved this manuscript.

\bibliographystyle{ieeetr}
\bibliography{lost}

\end{document}